\documentclass[conference]{IEEEtran}
\IEEEoverridecommandlockouts
\usepackage{cite}
\usepackage{amsmath,amssymb,amsfonts}
\usepackage{algorithmic}
\usepackage{graphicx}
\usepackage{textcomp}
\usepackage[table,xcdraw]{xcolor}

\usepackage[colorinlistoftodos]{todonotes}
\usepackage{float}
\usepackage[caption = false]{subfig}
\usepackage{amssymb}
\usepackage{graphicx}
\usepackage{cite}
\usepackage{pdflscape}
\usepackage{booktabs}
\usepackage[normalem]{ulem}
\usepackage{color,soul}
\usepackage[colorinlistoftodos]{todonotes}

\usepackage{algorithmic}
\graphicspath{ {Images/} }
\usepackage{hyperref}

\usepackage{array}
\usepackage{multirow}

\def\BibTeX{{\rm B\kern-.05em{\sc i\kern-.025em b}\kern-.08em
    T\kern-.1667em\lower.7ex\hbox{E}\kern-.125emX}}
\begin{document}

% \GDH{Paper title suggestion: Crossing the reality gap via simulator tuning for Manipulation.  Simulator tuning for sim2real. }
% \juxi{Traversing the reality gap ;)}
% \JC{It was originally "Traversing the Reality Gap through Simulator Tuning", back to this then.}

\title{Traversing the Reality Gap via Simulator Tuning\\
\author{Jack Collins$^{1,2}$, Ross Brown$^{2}$, J\"urgen Leitner$^{2,3,4}$ and David Howard$^{1}$%
\thanks{This research was supported by a Data61 PhD Scholarship.}% <-this % stops a space
\thanks{$^{1}$ Data61/CSIRO, Brisbane, Australia}%
\thanks{$^{2}$ Queensland University of Technology (QUT), Brisbane, Australia}%
\thanks{$^{3}$ Australian Centre for Robotic Vision (ACRV)}%
\thanks{$^{4}$ LYRO Robotics Pty Ltd, Brisbane, Australia}%
}
}

\maketitle

\begin{abstract}
The large demand for simulated data has made the reality gap a problem on the forefront of robotics. We propose a method to traverse the gap by tuning available simulation parameters. Through the optimisation of physics engine parameters, we show that we are able to narrow the gap between simulated solutions and a real world dataset, and thus allow more ready transfer of leaned behaviours between the two. We subsequently gain understanding as to the importance of specific simulator parameters, which is of broad interest to the robotic machine learning community. We find that even optimised for different tasks that different physics engine perform better in certain scenarios and that friction and maximum actuator velocity are tightly bounded parameters that greatly impact the transference of simulated solutions.
\end{abstract}

\begin{IEEEkeywords}
Reality Gap, sim2real, Differential Evolution, Simulator
\end{IEEEkeywords}

\section{Introduction}
Physics simulations attempt to model some pertinent facets of the real world in software. Simulations are necessarily simplified for computational feasibility, yet reflect real-world phenomena at a given level of veracity, the extent of which is the result of a trade-off between accuracy and computational time. In the domain of robotics, rigid-body simulators are frequently used as a large proportion of robots can be well-modelled as rigid bodies. Robotics simulators reproduce the most important physical phenomena (i.e. gravity, collisions, etc.) but replace detailed modelling of complex phenomena with computationally faster, less accurate high-level representations and constraints. 
Robotics simulations variously rely on the replication of phenomena that are difficult to accurately replicate,  e.g., simulating actuators (i.e.~torque characteristics, gear backlash, ...), sensors (i.e.~noise, latency, ...), and rendered images (i.e.~reflections, refraction, textures, ...). This gap between reality and simulation is commonly referred to as the ``Reality Gap''.

Although conducting research in simulation means having to overcome the reality gap, the associated pros outweigh the cons for many learning-based approaches. In simulation there is no risk of damaging hardware whilst having access to robots that are not physically available.  In addition, many instantiations of a scene can be run in parallel potentially faster than real-time, and human intervention is not required to manage experiments. With the current surge in data-driven techniques like Deep Learning, simulation data is either a pre-requisite to using such techniques, or at least a more attainable alternative to the (generally) expensive, laborious and non-scaleable collection of real-world data.

\begin{figure}[t]
	\centering
	\includegraphics[width=\linewidth]{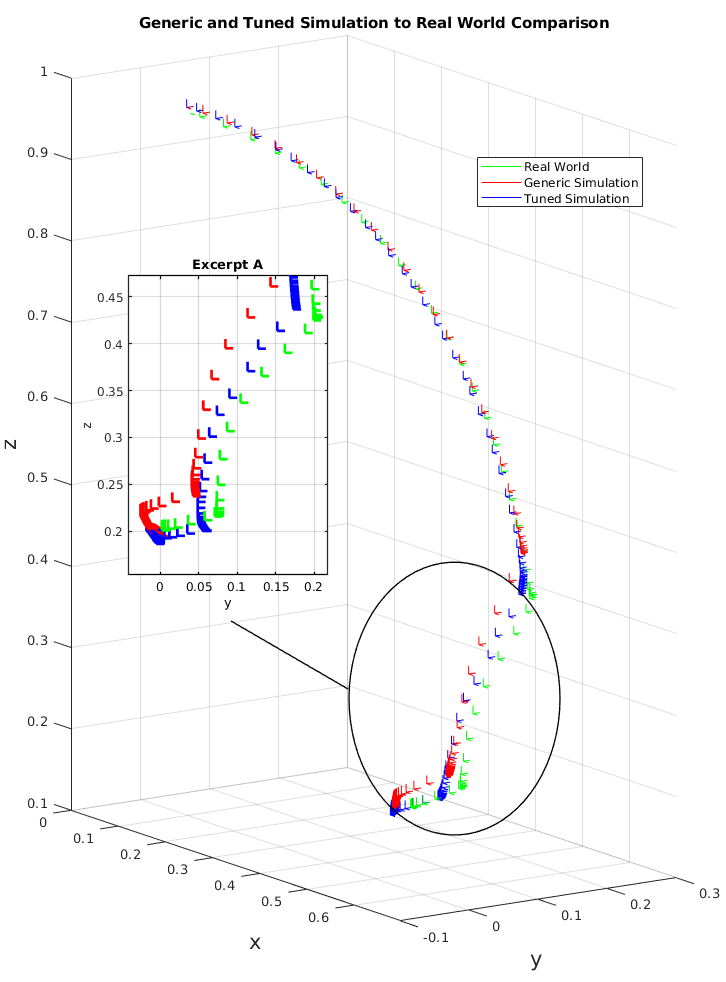}
	\caption{3D plot of a robotic manipulators end effector trajectory comparing (i) real world, (ii) untuned simulator, and (iii) tuned simulator. See magnified Excerpt A which visualises the advantages of the tuned simulation over the generic simulation.}
	\label{TitleImage}
	\vspace{0mm}%-4mm}
\end{figure}

Of course, working in simulation only makes sense if the eventually-learned behaviour can transfer to reality. {\em Sim2real} research aspires to make any simulated behaviour seamlessly transfer to hardware and operate in real-world conditions. There are three prevalent ideologies within the sim2real community for overcome the reality gap, (i) data-driven improvement of simulation, (ii) generation of robust controllers, and (iii) a hybrid approach combining (i) and (ii). (i) augments the simulation with real-world data. This approach suffers as collection of data from the real world remains expensive. In comparison, (ii) must expose a controller to a wide range of environments through the randomisation of a subset of simulation parameters or introduction of noise. Due to the controller being exposed to both realistic and (a large number of) unrealistic scenarios the creation of such controllers is time consuming. (iii) attempts to mitigate the disadvantages of collecting real world data by hand and simulating a large number of unrealistic training scenes but in doing so adds the complexity of integrating simulation and the real world into a single workflow.

% \GDH{ I  feel like this  paper  is more about demonstrating a technique that builds on our previously-RAL  accepted benchmark.  So we have identified parameters that make sense to benchmark, and we have  devised a protocol to capture them.  Now we  are using that to improve the transfer of controllers from simuulation to reality, by first  tuning  the simulator towards our  real robot  environment... We are -also- showing researchers which parameters matter more, or induce the most variance  in simulator perforrmance over which  ranges.}

In this work we build on past research that describes a method for recording data for comparison between reality and simulation \cite{Collins2019QuantifyingTasks}, as well as the provision of metrics for a principled, numerical quantification of the differences between simulation and reality. Our extension investigates the optimisation of simulation parameters towards the goal of achieving real world simulation performance. We show that our approach is able to attain better performance than generic simulator parameters (as visualised in Figure~\ref{TitleImage}) and is a promising method for traversing the reality gap. Our contributions aside from a method to optimise simulator parameters include an analysis as to the most influential parameters in achieving real world results.

% \GDH{WHAT DO WE DO IN THIS PAPER?}
% \GDH{WHAT ARE THE NOVEL CONTRIBUTIONS?}

% \GDH{The -MAIN- motivation is to show this as a good approach for doing sim2real! The -SECONDARY- motivation is to inform researchers as to the importance of various simulator parameters?} 

The motivation for our work is to inform researchers and those applying sim2real techniques as to the important parameters to tune in simulation. As an extension of the results, further conclusions can be drawn as to the best data to collect from the real world to make realistic simulations, the parameters to randomise and the extent of learning approach refinement.

The question we endeavour to answer is; \textit{what are the simulation parameters that are most influential in arriving at a realistic simulation?} Our experimentation involves optimising two popular rigid body simulators used by the robotics community to faithfully replicate the results of a series of tasks conducted by a real robot in a motion capture system. The tasks are a range of kinematic movements and object interactions performed by a robotic manipulator, and the results of this `ground truth' are available as a publicly-accessible dataset \cite{Collins2019BenchmarkingDataset}. Optimisation of simulator parameters is via differential evolution, an Evolutionary Algorithm that performs well on high-dimensional optimisation problems.

\section{Background}

\subsection{Simulation and Physics Engines}
Rigid-body simulators are a class of simulators that simplify the world into rigid objects that are potentially connected through (actuated or unactuated) joints. The use of rigid-body simulators is prevalent in robotics due to the widespread use of rigid robots in the field. Simulators are often modular, with the simulator acting as a high level interface, typically including a graphical user interface, API accessibility from external programming languages, plugins, importers, and scene description formats \cite{Torres-Torriti2016SurveyRobots}.  

Robotic simulators utilise one or more of a number of physics engines. Common physics engines include Bullet \cite{E.CoumansandY.Bai2016PybulletLearning}, Dynamic Animation and Robotics Toolkit (DART) \cite{Lee2018DART:Toolkit} and Open Dynamics Engine (ODE) \cite{Smith2005OpenEngine}, all of which are licensed under free software licenses. The physics engine operates below the simulator with the goal of providing physically accurate movement of objects instantiated in the simulator. 

A multitude of user definable parameters are available to be tuned, however the exact number varies between engines and implementations. Parameters relate to the visual aspects (colours, textures, etc. ), material properties (frictions, restitution, etc.), object properties (mass, inertia, etc.), joint characteristics (type of constraint, actuation, etc.) and other more general physics engine properties (time step, solver settings, dampening, etc.).  Although a large tunable parameter space represents an opportunity to adapt a physics engine to accurately replicate real-world conditions, it also results in a high-dimensional optimisation problem, where the effects of varying parts of this parameter space on simulator performance is not intuitive.

\subsection{Reality Gap}
% \GDH{brief intro on e.g. tune the controller to cross, select a controller that can cross (Koos 2010), tune the sim (pick a reference), add noise (jakobi), learn detailed subsystem dynamics (Hwangbo)}
With the well accepted problem of the reality gap the only way currently to guarantee a solution will perform as expected in reality is to create the solution in reality, to this end test-rigs are a proven method for overcoming some of the issues associated with working in the real world \cite{Howard2015AControllers,Heijnen2017AHardware}. There are a range of approaches that have been raised in the past to cross the reality gap starting most notably with Jakobi et al. \cite{Jakobi1995NoiseRobotics} using targeted noise to generate robust controllers. Other sim2real approaches have focused on tuning controllers~\cite{qiu2020crossing}, optimising transferable controllers \cite{Koos2010CrossingControllers}, adding perturbations to the environment \cite{Tan2018Sim-to-Real:Robots} and learning the target platforms actuator responses \cite{Hwangbo2019LearningRobots}.

There are several methods of overcoming the reality gap using real-world data augmentation. The most common is to alter the generic simulator settings with more accurate parameters that are collected from real world measurements, derived from calculations, from researched values or experimentally \cite{Tan2018Sim-to-Real:Robots, Tan2016Simulation-basedBalancing, Williams2002DynamicRobots}. Oftentimes parameters such as weight, physical dimensions, frictional coefficients, centre of mass, inertial properties, actuator control properties and more are used \cite{Gautier1988OnRobots, Chebotar2019ClosingExperience}. Another common practice is to substitute a more accurate model of an actuator derived from the response of the physical robot or parameters from system identification, examples of this include work by Andrychowicz et al. \cite{Andrychowicz2018LearningManipulation} and Tan et al. \cite{Tan2018Sim-to-Real:Robots}.

To update parameters using recorded data there are several documented approaches with a portion of these updating simulation parameters live. One example of live parameter updates is by Moeckel et al. using a Kinect sensor coupled with background subtraction \cite{Moeckel2013GaitReality} to detect gait transference. Although motion capture systems that give accurate 6DOF pose are becoming increasingly common equipment in research labs, few methods have been reported that use them \cite{Andrychowicz2018LearningManipulation}.

Domain randomisation is currently the most popular method for overcoming the reality gap in the machine learning domain. The parameters to be randomised are chosen according to the policy which is being learned, tasks that are entirely visual or require computer vision focus on randomising the visual characteristics of a simulation (i.e. parameters relating to cameras, colours, textures, lighting, etc.) \cite{Tremblay2018TrainingRandomization}. A policy that requires environmental interactions will randomise physical properties (i.e. mass, inertia, friction, etc. ) \cite{Peng2018Sim-to-realRandomization}. Randomised parameters are bounded with limits at initialisation which are hand picked by the user, these are often plausible ranges that will be found within the real world, although not necessarily accurate to the operating conditions of the target robot and environment. 

By presenting such varied scenes to an agent, extensive amounts of time are required to learn. In particular, slow initial learning rates are noted, and approaches have attempted to overcome this by progressively increasing the variation experienced \cite{OpenAI2019SolvingHand}. Accurate parameter ranges that are specific to the parameter settings would further reduce the landscape and lead to quicker training times, and this is one contribution of our work.

\subsection{Parameter Optimisation}

As discussed there are a large number of available parameters to optimise when applied to simulators and physics engines. As rigid-body simulators commonly used in robotics are non-differentiable \cite{Degrave2019ARobotics} the optimisation of parameters relies on gradient free algorithms. Bayesian optimisation was considered for the task of finding a global minimum as it provides an efficient sampling method requiring reduced simulation evaluations \cite{Snoek2012PracticalAlgorithms}. However, Bayesian optimisation is limited in the number of variables it is able to optimise, common practice is to optimise up to $10$ \cite{Wang2013BayesianEmbeddings}. Extensions of Bayesian Optimisation have seen this extended further up to $30$ variables using drop out \cite{Li2018HighDropout}.

Evolutionary Algorithms (EAs) are another class of black-box optimisation algorithms that have been applied to problems with larger search spaces.  Differential Evolution (DE)\cite{Storn1997DifferentialSpaces} is a popular EA, a global optimiser this is easily parallelisable and able to scale to a large number of variables. Algorithmic parameters for tuning include crossover rate $CR$, mutation weight $F$ and population size $N$ \cite{Das2011DifferentialState-of-the-Art}. A good value for $CR$ has been found to be between $0.3$ and $0.9$ according to Ronkkonen et al. \cite{Ronkkonen2005Real-parameterEvolution}. The population size is closely related to the mutation parameters, with problem dimensionality and problem properties also affecting the choice in population size. The recommended population size for $30-50$ dimension problems is $3d-5d$ from a review conducted by Piotrowski \cite{Piotrowski2017ReviewSize}.

\section{Methodology}

Our approach to simulator optimisation can be segmented into several components as listed below:

\begin{itemize}
    \item Real World Dataset Collection (Section~\ref{dataset})
    \item Robotic Simulations (Section~\ref{simulation})
    \item Simulator Parameter Selection (Section~\ref{parameters})
    \item Running the Optimisation Algorithm (Section~\ref{optimisation})
\end{itemize}

\subsection{Dataset} \label{dataset}

Our dataset is a publicly available collection of tasks completed by a robotic manipulator and recorded by a motion capture system \cite{Collins2019BenchmarkingDataset}. The data gives a ground truth of the real world with 6DOF pose of the manipulator and manipulated objects recorded. There are 10 tasks in total, 2 of which are pure kinematics (no objects) and 8 of which involve non-prehensile manipulation. Tasks are purposefully elementary as they are foundational to larger compound tasks, making results derived from these tasks scale to harder and more complex applications. The manipulation tasks have interactions with objects including cubes, cuboids, cylinders and cones. Another useful property of the dataset is the contrast in object materials, with half the interaction tasks completed with plastic objects and the other half with exact wooden replicas. For a complete description of the dataset and its use in benchmarking reality, please see \cite{Collins2019BenchmarkingDataset}.

The dataset is released with simulation protocols that allow users to simulate the same scene and same control of the robot arm that is used in reality. All scenes use a levelled plane with a Kinova 6DOF robotic arm attached with KG-3 gripper and either none or one object to manipulate. Dataset users must follow the explicit instructions on scene setup, robot configuration and motor controls, but are able to change any of the other user definable parameters of their chosen simulator and physics engine. 

\subsection{Simulation} \label{simulation}

We selected two popular robotic simulators; PyBullet (version 2.5.8) \cite{E.CoumansandY.Bai2016PybulletLearning} and V-Rep (version 3.6.2 now known as CoppeliaSim) \cite{Rohmer2013V-REP:Framework}, accessed through the PyRep interface \cite{James2019PyRep:Learning}. The two simulators are chosen as they provide a common interface, and also provide easy access to a multitude of physics engines. Pybullet uses Bullet 2.85 whilst V-Rep uses Bullet 2.83 and Bullet 2.78. V-Rep also provides access to ODE and Newton physics. 

PyBullet exposes a large number of settings to the user natively. V-Rep has an abstraction layer between the simulator and physics engines making it possible to interface with multiple different engines. Most of the same parameters accessible to the PyBullet interface are available in the V-Rep physics engines. However, as we use the PyRep interface not all parameters we require are accessible from the external API therefore we use embedded scripts within the simulator which are invoked from PyRep. 

\subsection{Parameters} \label{parameters}

From the $5$ different physics engines available (including the 3 versions of Bullet) there are many parameters that create the same effect on the physics of the simulation that are either implemented using different methods, or different units. As such it was necessary to find the shared parameters that were directly comparable between physics engines, and settings that were not. We used two approaches; in the first we compared only those parameters available across all simulations and physics engines (Shared).  In the second we allow each to tune a fuller range of parameters that may be available (Individual).  Table \ref{Parameters} documents all the Shared parameters and the Individual parameters that we chose to simulate. The Individual parameter optimisation included both the Shared and Individual parameters. 

% \usepackage{graphicx}
\begin{table*}[t]
\caption{List of Parameters Used for Optimisation and the Range Limits}
\label{Parameters}
\resizebox{\textwidth}{!}{%
\begin{tabular}{|l|l|l|l|}
\hline
\textbf{Shared} & \textbf{Range} & \textbf{Individual} & \textbf{Range} \\ \hline
Time Step & {[}0.001,0.05{]} & Joint Damping (6 Joints) & {[}0.0001,0.9{]} \\ \hline
Mass (Links, Gripper, Objects) & {[}0.7*M,1.3*M{]} & Rolling Friction (Gripper,Floor, Objects) & {[}0.0001, 1.25{]} \\ \hline
Maximum Joint Torque (6 Joints) & {[}100,9000{]} & Sliding Friction (Gripper,Floor, Objects) & {[}0.0001, 1.25{]} \\ \hline
Maximum Joint Velocity (6 Joints) & {[}10,40{]} & Restitution (Gripper,Floor, Objects) & {[}0.0001,0.9{]} \\ \hline
Lateral Friction (Gripper,Floor, Objects) & {[}0.0001, 1.25{]} & Linear Damping (Gripper,Floor, Objects) & {[}0.0001,0.9{]} \\ \hline
 &  & Angular Damping (Gripper,Floor, Objects) & {[}0.0001,0.9{]} \\ \hline
\end{tabular}%
}
\end{table*}

It would be infeasible to tune all available parameters. As positions (x,y,z), rotations (x,y,z,w) and inertias (xx,yy,zz) require multiple parameters each, it was impractical to create variables for the centre of masses, inertia position, inertia rotation and inertia tensor. %In addition, other parameters were disregarded if they did not have a shared counterpart with any other physics engine. 

\subsection{Optimisation} \label{optimisation}

We use DE as implemented in the SciPy optimise module. DE follows the same approach as most EAs in that it begins with a randomly initialised \emph{population} of a set number of individuals evolved across a number of \emph{generations}. Individuals are a vector of parameters and child populations are the succeeding generation (or offspring) from parent populations with a chosen strategy dictating the creation of child populations. 

We apply 'best1bin' strategy which iterates over the parent population creating a vector  for each individual ($X_{i}'$) by mutating the fittest individual in the parent population ($X_{best}$) by the difference between two randomly chosen individuals of the parent population, see Equation~\ref{DE} where $F$ is the mutation factor. A child member is then created by choosing each parameter from either $X_{i}'$ or the $i^{th}$ parent as per a binomial distribution where the number must be less than the recombination rate to select the parameter from $X_{i}'$. If a child vector is fitter than its parent it replaces it in the current population. In comparison to other strategies 'best1bin' has strong supporting evidence that it is a competitive strategy \cite{Mezura-Montes2006AOptimization}.

\begin{equation}
\label{DE}
X_{i}' = X_{best} + F \times (X_{1\_rand}+X_{2\_rand})
\end{equation}

The fitness objective is to minimise the 3D Euclidean distance between the simulator and reality, this value dictates the fitness used by the DE for a given population member. For tasks $1$ and $2$ (kinematic tasks) this is the distance between the wrist joint of the robotic manipulator in simulation ($W_{x,y,z}^{s}$) and the dataset ($W_{x,y,z}^{d}$) summed at $20Hz$ throughout the duration of the simulation and divided by the number of data points ($n_{points}$), see Equation~\ref{euclideanError}. Tasks that include objects (simulation: $O_{x,y,z}^{s}$, dataset: $O_{x,y,z}^{d}$) use the combined euclidean distance of the arm and object, see Equation~\ref{combinedError}. The trajectory of the dataset object is a distribution as you can not create a mean object trajectory from multiple repeats where the object did not have the same start and end position. Therefore, we use the difference in the final position of the object in simulation and the dataset.

\begin{equation}
f =  \displaystyle\frac{\sum_{points}\sqrt[]{\sum_{i=x,y,z} (W_{i}^{d}-W_{i}^{s})^{2}}}{n_{points}}
\label{euclideanError}
\end{equation}

\begin{equation}
f =  \displaystyle\frac{\sum_{points}\sqrt[]{\sum_{i=x,y,z} (W_{i}^{d}-W_{i}^{s})^{2}}}{n_{points}} + \sqrt[]{\sum_{i=x,y,z} (O_{i}^{d}-O_{i}^{s})^{2}}
\label{combinedError}
\end{equation}

\section{Experimentation}

We define an optimisation as an application of DE to optimise an array of either shared or individual simulation parameters. In total there were $1100$ optimisations completed that make up the results. This is broken down into:

\begin{itemize}
    \item ``Shared'' and ``Individual'' parameters;
    \item $11$ Experiments: $10$ manipulation tasks from the dataset the $11^{th}$ a combination of all $10$;
    \item 5 physics engines; and
    \item 10 repeats of each.
\end{itemize}

A large number of experimental runs were required and as such were scheduled on a High Performance CPU Cluster (HPC). A singularity container with PyBullet, PyRep and V-Rep installed provided a distributed and scaleable deployment. Experiments $1-11$ were paired and scheduled onto a single node of the HPC, with each repeat given access to a single core. Each node was 20 cores of an Intel Xeon E5-2660 V3 processor with a clock speed of $2.6 GHz$, $25MB$ cache and $128GB$ of memory. 

The constants for the DE algorithm were $0.7$ for $CR$, $0.5-1.0$ for $F$ with dithering and a population of $N = 1D$ where $D$ is the length of the parameter array. The population was chosen to be low due to the large evaluation times of some physics engines and experiments paired with the limited amount of compute time. The experiment finished under one of three conditions:
\begin{itemize}
    \item Convergence (i.e. when the standard deviation of the current population was less than one percent of the population mean), or
    \item $1000$ DE generations completed, or
    \item $168hrs$ of compute time (a hard constraint of the HPC).
\end{itemize}

The number of variables tuned varied for Shared and Individual experiments. Shared experiments tuned $31$ variables, Individual experiments tuned $57$ (inclusive of the Shared variables). See Table \ref{Parameters} for more details.

\section{Results}

% Using a Mann Whitney U-Test to compare the optimised fitnesses to the fitnesses achieved by generic simulator parameters we see in Table~\ref{mannwhitney} that all but three experiments optimising individual parameters are statistically significant ($\rho<0.05$). This is statistically significant evidence that tuning the simulation . When performing the same statistical test between shared and individual parameters we see that only $25$ of the $55$ ($11$ Experiments $\times \: 5$ Physics Engines) produce statistically significant improvements with no clear trend to comment on.  

\subsection{Shared Parameters}

\subsubsection{Performance}
\begin{figure}[t]
	\centering
	\includegraphics[width=\linewidth]{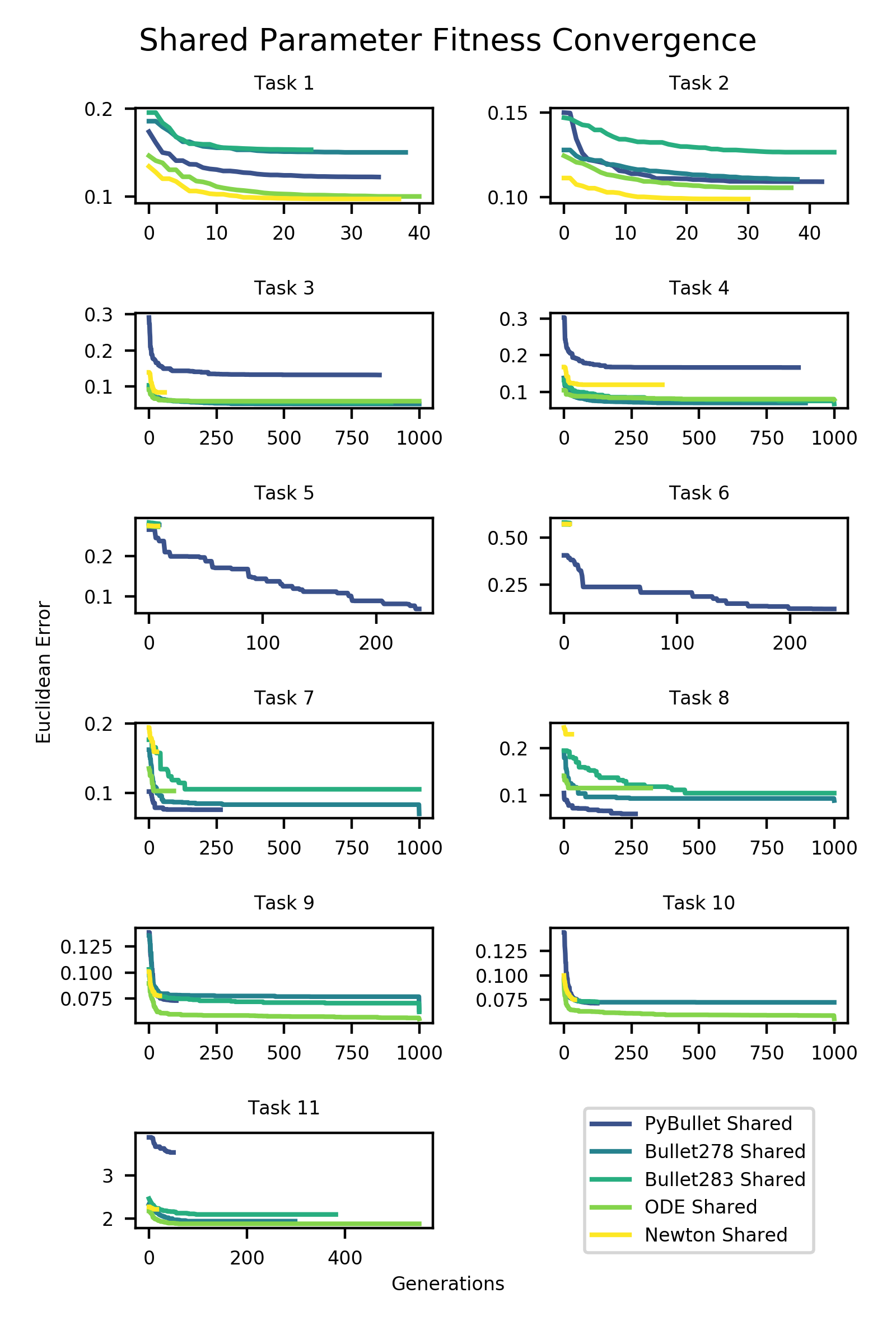}
	\caption{Convergence plots of all $11$ experiments using shared parameters. Each subplot plots the line of best fitness throughout the generations averaged across the $10$ repeats for each physics engine.}
	\label{Shared_Plot}
	\vspace{0mm}%-4mm}
\end{figure}

Table \ref{genericFitness} shows the fitnesses for each experiment and physics engine when using generic simulator parameters. The fitness plots in Figure \ref{Shared_Plot} demonstrate the convergence of the optimisations completed on the same physics engines, and the parameters shared between all physics engines. The smallest error obtained between simulation and the real world dataset for each experiment can be found in Table \ref{lowestError}.

Directly comparing the generic fitness values from Table \ref{genericFitness} to the fitness values achieved when optimising shared parameters in Table \ref{lowestError} we see that the tuned fitness is lower than all physics engines with generic parameters. Taking the best generic fitness and comparing it to the best tuned fitness for each experiment we see improvements ranging from $14\%$ for experiment $2$ and $91\%$ for experiment $6$. The effect of tuning the shared parameters is therefore significant and provides a more realistic simulation closer aligned to the real world.

From Table \ref{lowestError} we see a correlation between the experiment 'type' and the physics engine with the least error. Newton was the best Physics Engine for Experiments $1$ and $2$, implying that it is able to better model arm kinematics without object interactions. PyBullet was best at experiments $5,6$ and $8$, all of which include rolling objects. The clustering of experiments and physics engines indicates that no physics engine is best equipped to deal with all simulation scenarios but that physics engines can have heightened performance in select scenarios over other physics engines.

% \usepackage{graphicx}
\begin{table}[]
\caption{Fitness of Generic Physics Engine Settings}
\label{genericFitness}
\resizebox{\linewidth}{!}{%
\begin{tabular}{|c|l|l|l|l|l|}
\hline
\multicolumn{1}{|l|}{\textbf{Experiment}} & \textbf{PyBullet} & \textbf{Bullet2.78} & \textbf{Bullet2.83} & \textbf{ODE} & \textbf{Newton} \\ \hline
1 & 0.1498 & 0.2660 & 0.2687 & 0.1800 & 0.1437 \\ \hline
2 & 0.1283 & 0.1320 & 0.1582 & 0.1285 & 0.1147 \\ \hline
3 & 0.1764 & 0.2533 & 0.2838 & 0.1943 & 0.2691 \\ \hline
4 & 0.2107 & 0.2255 & 0.1966 & 0.1674 & 0.2877 \\ \hline
5 & 491.2769 & 0.2967 & 0.3031 & 0.2909 & 0.2910 \\ \hline
6 & 503.3037 & 0.6029 & 0.6096 & 0.5972 & 0.5977 \\ \hline
7 & 0.1778 & 0.2500 & 0.3163 & 0.2370 & 0.2323 \\ \hline
8 & 0.2132 & 0.3075 & 0.3412 & 0.2812 & 0.2759 \\ \hline
9 & 0.1306 & 0.1950 & 0.1241 & 0.1171 & 0.1242 \\ \hline
10 & 0.1242 & 0.1143 & 0.1155 & 0.1069 & 0.1176 \\ \hline
11 & 995.8916 & 2.6433 & 2.7170 & 2.3005 & 2.4540 \\ \hline
\end{tabular}%
}
\end{table}

\begin{table}[]
\caption{Best Optimised Fitness by Experiment and Parameters Tuned}
\label{lowestError}
\resizebox{\linewidth}{!}{%
\begin{tabular}{|c|l|l|}
\hline
\multicolumn{1}{|l|}{\textbf{Experiment}} & \textbf{Best Physics Engine Shared} & \textbf{Best Physics Engine Individual} \\ \hline
1 & Newton (0.0973) & Newton (0.0973) \\ \hline
2 & Newton (0.0984) & Newton (0.0984) \\ \hline
3 & Bullet 2.78 (0.0498) & Bullet 2.78 (0.0498) \\ \hline
4 & Bullet 2.83 (0.0629) & Bullet 2.78 (0.0673) \\ \hline
5 & PyBullet (0.0506) & PyBullet (0.2407) \\ \hline
6 & PyBullet (0.0552) & PyBullet (0.5641) \\ \hline
7 & Bullet 2.78 (0.0551) & PyBullet (0.0551) \\ \hline
8 & PyBullet (0.0442) & Bullet 2.78 (0.0744) \\ \hline
9 & ODE (0.0519) & ODE (0.0482) \\ \hline
10 & ODE (0.0503) & ODE (0.0487) \\ \hline
11 & ODE (1.7360) & ODE (1.7714) \\ \hline
\end{tabular}%
}
\end{table}

\begin{figure}[t]
	\centering
	\includegraphics[width=\linewidth]{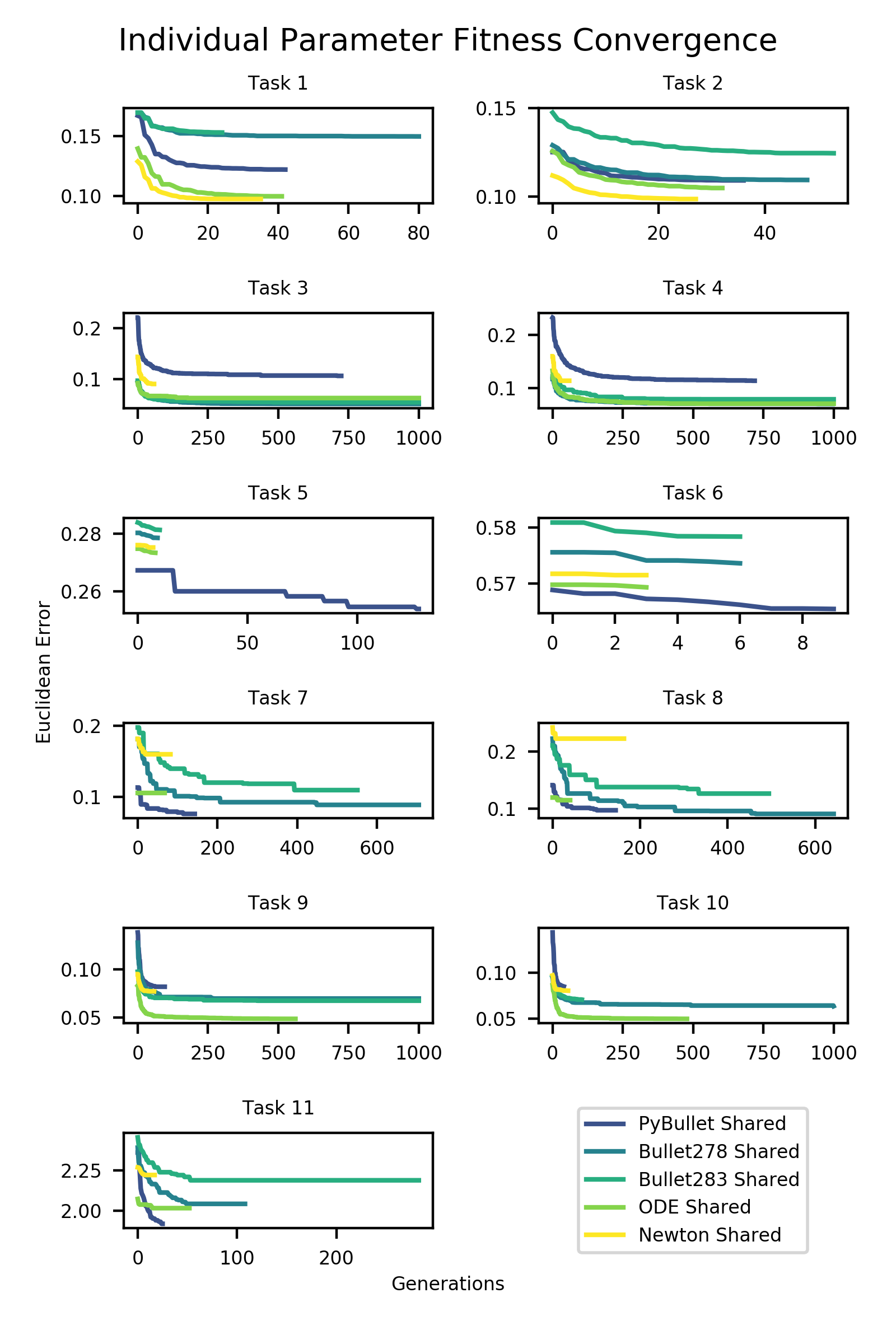}
	\caption{Convergence plots of all $11$ experiments using individual parameters. Each subplot plots the line of best fitness throughout the generations averaged across the $10$ repeats for each physics engine.}
	\label{Indiviual_Plot}
	\vspace{0mm}%-4mm}
\end{figure}

Newton took the least compute time, taking a total $279hrs$ for all $11$ experiments. Slowest was PyBullet at $662hrs$. The time required to complete an optimisation gives some notion of the difficulty, as experiment 11 (the combination of all dataset tasks, i.e. $1-10$) understandably took the longest. We note extended completion times for experiments $7$ and $8$, which were cylinder rolling tasks. 

Table \ref{Shared_Plot} displays in the y-axis of each subplot the number of generations required for the optimisation to terminate. Newton was consistently the physics engine with the lowest number of evolutionary generations, whilst Bullet 2.83 and Bullet 2.78 had the most, terminating at $1000$ generations for experiments $9$ and $10$ both of which were comparatively easy experiments. This is an interesting observation as it alludes to the fact that Newton is an easier environment to optimise within, with either a reduced search space or less noise, however, Newton does not provide accurate environmental interactions when compared with the other physics engines being reviewed. The large number of evolutionary generations required by both of V-Reps Bullet environments especially for the easier cuboid interactions implies a noisy landscape for fitness optimisation.

\subsubsection{Parameters}

\begin{figure}[htb]
	\centering
	\includegraphics[height=0.75\textheight]{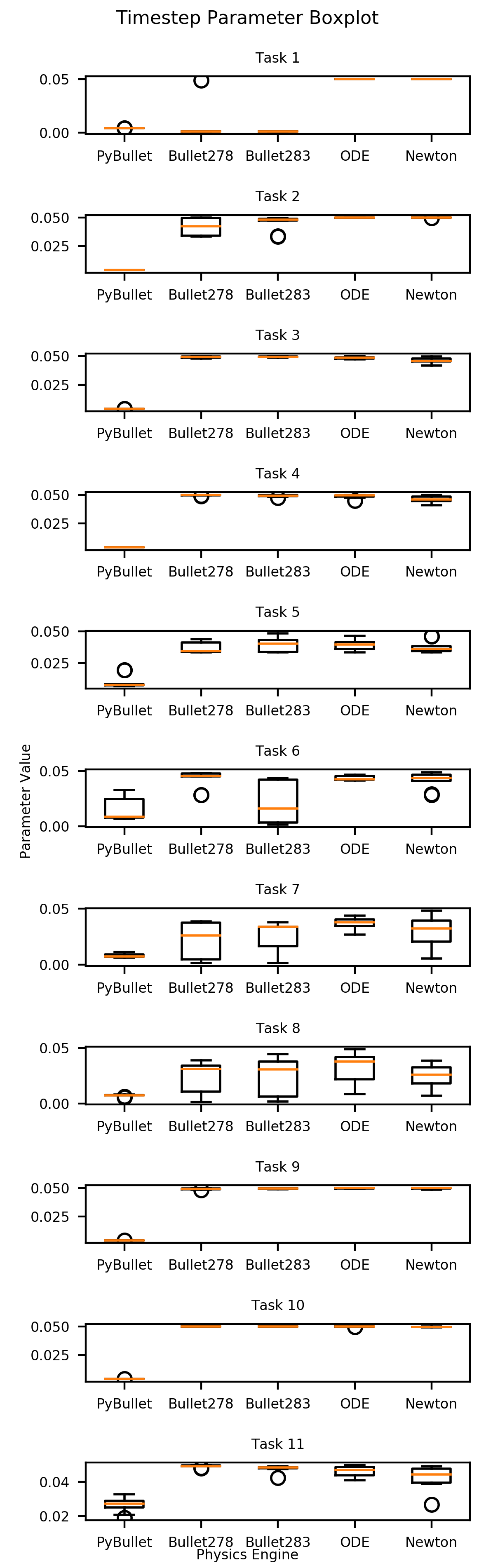}
	\caption{Box and Whisker Plot for each (i) physics engine and (ii) experiment. The subplots are all for the shared timestep parameter. This plot is meant as a exemplar of the other $30$ parameter plots for individual optimisations and $56$ parameter plots for shared optimisations.}
	\label{timestep}
	\vspace{0mm}%-4mm}
\end{figure}

From the Shared parameters we note a select few having a large impact on the performance of the Physics engine. We measure the importance of a parameters value as its deviation across the 10 optimisation repeats. Owing to the large amount of data generated, we include exemplar box and whisker plots in Figure \ref{timestep} for most relevant data.

One of the most influential parameters found was the simulation timestep (see Figure \ref{timestep}), the deviation was consistently low across experiments and physics engines except for rolling tasks (experiments $5-8$). The generic timestep value for V-Rep is $0.05sec$ and $0.0041sec$ for PyBullet. Pybullet's median value was $0.0042sec$ with a standard deviation of less than $0.0095sec$ for all experiments. Similarly, V-Rep Physics Engines were also very close to the generic timestep with the median for Bullet 2.78: $0.4304sec$, Bullet 2.83: $0.0456sec$, ODE: $0.0485sec$ and Newton: $0.0459sec$. We therefore recommend setting the physics engine timestep to the recommended value as detailed by the developer of the simulator. The constrained value is very likely to be due to a reliance of other parameters that would need to be tuned that are physics engine specific. 

Other parameters that largely influenced the realism of the simulation were the lateral friction of the manipulated objects i.e. wood and plastic frictions. The friction of the gripper and floor plane were not as influential except for Pybullet during the two rolling experiments where the floor plane had a standard deviation of $0.0097$ and $0.1465$. This is a very likely the reason why PyBullet had the lowest error for three of the rolling experiments.  

It was expected that the parameters influencing the response of the joints would have a large impact on the fitness as there is a direct correlation between the measured wrist joint and actuator response. This assumption was found to be true for the maximum joint velocity, but no such trends could be found for the maximum torque. Experiments $1-4$ consistently had statistically significant lower standard deviation across joints $1-5$. Joint $6$ was likely less influential in simulator realism due to the restricted amount of movement it experienced in the experiments and although experiments $5-10$ did not display the same reliance on accurate joint velocity this is likely due to the experiments being more complex and the resulting optimisation harder. The results from the shared parameter optimisation show that we can perform context-sensitive tuning that is able to positively influence the realism of the simulation for all environments.

\subsection{Individual Parameters}

\subsubsection{Performance}

Figure \ref{Indiviual_Plot} depicts the fitness convergence for the $57$ individual parameters tuned. Made obvious by the plots is the difficulty that the extra parameters add as some physics engines fail to converge appropriately for certain experiments. When comparing the lowest error for each experiment as found in Table \ref{lowestError}, there are only three instances where the individual optimisation improves upon the shared parameters and $5$ where the optimisation arrives at a worse solution. The added complexity of the additional parameters to tune is likely the cause of the worse fitnesses. Similar to the shared parameter optimisations the individual runs have a correlation between the best physics engine and the type of task, i.e. Newton is best at kinematics and PyBullet is best at $3$ of the $4$ tasks that include rolling objects.

Taking into account the mean final fitness instead of the absolute best the individual optimisation lessens the engine/experiment error $35$ out of the $55$ times. This is likely due to several reasons, (i) DE does not guarantee to find the optimal solution, (ii) the additional parameters add noise and complexity that make it harder to optimise, and (iii) due to termination of compute after $168hrs$ the algorithm is unable to complete optimisation without convergence. Termination at the maximum run time of individual parameter optimisations occurs on $14$ out of $55$ occasions.

% \GDH{We cannot conclude that Individual is better or worse, need access to a different HPC...?}
\subsubsection{Parameters}

The influential parameters for individual optimisations were the same as those for shared. These being timestep, lateral friction and maximum joint velocity. In addition to these parameters there were two additional ones that had statistically significant standard deviations. The restitution of an object parameterizes the conservation of energy after a contact, only $9$ of our $11$ experiments contain contacts, most of which are at low speed. Experiment $9$, interaction with a plastic cuboid, saw a standard deviation of less than $0.16$ for three of the physics engines for the restitution value of plastic. This eludes to the fact that restitution could be an important factor given experiments that include contacts that are above a contact energy threshold. 

The mass of arm links was another parameter that saw noticeably low standard deviations for individual parameter optimisations. Arm link masses produced an interesting result with links $1, 3, 4 and 5$ displaying low standard deviations for Bullet 2.78, Bullet 2.83 and ODE on the simpler experiments i.e. experiments $1-4$. The smaller deviations across the easier experiments is likely due to the reduced noise whilst optimising the parameters.

% -Fitness, completion times, etc.
%     -Plot of 11 experiments (in subplots) with 10 lines for each subplot (5 lines for each physics engine x2 becasue we will include shared and individual results on each plot)
%     -Box and Whisker of final euclidean distance error for all experiments,engines,shared/individual
% -Shared Params
%     -How many Plots???
%     -Look at those that were important to get correct (small spread/std. dev)
%         -Friction coefficients
%         -Simulator step size, why is this interesting? It is believed that reducing the step size increases the accuracy but this is not the case with the chosen robotic simulators
%         -Max velocity - task dependant??
        
%     -Additional (Yet to look indepth)
% -Individual Params
%     -Step size and frictions as above
%     -Additional (Yet to look indepth)

\section{Conclusion}

In conclusion, we have investigated the influence of a range of simulation parameters on optimising 2 simulators and 5 physics engines towards more realistic simulations. Our method is significantly better than using generic simulator parameters with all simulation environments and all experiments achieving an improved fitness. This was achieved by using a real world dataset of motion capture recorded manipulation tasks and optimising both shared and individualised simulation parameters towards the dataset. The optimisation algorithm chosen was differential evolution (DE) due to the large number of optimisation parameters it is able to concurrently optimise and the non-differentiable nature of the problem. The fitness signal throughout the optimisation runs was the Euclidean distance error between the simulated wrist of the manipulator summed with the final placement error of any objects in the scene. 

We found that the most important parameters that were shared between physics engines were simulation timestep, lateral object friction and joint velocity. From the expanded range of parameters we also found that it is likely for high energy contact tasks that the value of restitution is important. 

To improve simulator performance we recommend that users start with (i) the default simulator timestep, (ii) researching or experimentally acquiring an accurate friction value, and (iii) recording or sourcing accurate maximum joint velocities for each actuator. These same parameters, if accurate, should not be excessively randomised as results indicate that the variation in these parameters are tightly bounded to the real world value.

% \GDH{One issue we havent looked at is if the tuned params can balance out to give good fitness, but be physically impossible.. e.g. what happens if we tune gravity?  would be a good discussion or suggestion for future work}

% Finally, to prove that our solution is able to traverse the reality gap we transfer a controller generated in an optimised simulation (controller unseen from optimisation) to the real world. The task has a lower error when transferred than when transferred with optimised simulation parameters. 

% Conclude
% Recommend that leave the simulator step size as default, spend the time to calculate experimentally or from resources an accurate friction

% \section*{Acknowledgment}

% \section*{References}

\bibliographystyle{IEEEtran}
\bibliography{Mendley,additional}

\end{document}